# STAR: A Privacy-Preserving, Energy-Efficient Edge AI Framework for Human Activity Recognition via Wi-Fi CSI in Mobile and Pervasive Computing Environments

Kexing Liu

*Abstract*— Human Activity Recognition (HAR) via Wi-Fi Channel State Information (CSI) presents a privacy-preserving, contactless sensing approach suitable for smart homes, healthcare monitoring, and mobile IoT systems. However, existing methods often encounter computational inefficiency, high latency, and limited feasibility within resource-constrained, embedded mobile edge environments. This paper proposes STAR (Sensing Technology for Activity Recognition), an edge-AI-optimized framework that integrates a lightweight neural architecture, adaptive signal processing, and hardware-aware co-optimization to enable real-time, energy-efficient HAR on low-power embedded devices. STAR incorporates a streamlined Gated Recurrent Unit (GRU)-based recurrent neural network, reducing model parameters by 33% compared to conventional LSTM models while maintaining effective temporal modeling capability. A multi-stage pre-processing pipeline combining median filtering, 8th-order Butterworth low-pass filtering, and Empirical Mode Decomposition (EMD) is employed to denoise CSI amplitude data and extract spatial-temporal features. For on-device deployment, STAR is implemented on a Rockchip RV1126 processor equipped with an embedded Neural Processing Unit (NPU), interfaced with an ESP32-S3-based CSI acquisition module. Experimental results demonstrate a mean recognition accuracy of 93.52% across seven activity classes and 99.11% for human presence detection, utilizing a compact 97.6k-parameter model. INT8 quantized inference achieves a processing speed of 33 MHz with just 8% CPU utilization, delivering sixfold speed improvements over CPU-based execution. With sub-second response latency and low power consumption, the system ensures real-time, privacy-preserving HAR, offering a practical, scalable solution for mobile and pervasive computing environments.

*Index Terms*— Wi-Fi CSI, Human Activity Recognition, Edge AI, Privacy-Preserving Sensing, Embedded Edge Devices, Mobile IoT

## I. INTRODUCTION

In recent years, contactless human activity recognition (HAR) has emerged as a crucial technology for a wide range of applications, including smart homes [1, 2], health monitoring [3], intelligent security [4], and smart city infrastructure [5, 6]. These systems are gaining widespread adoption due to their non-intrusive nature, operational flexibility, and broad applicability in both residential and commercial environments, encompassing office spaces, public transportation systems, and densely populated public venues. In domestic settings, non-contact HAR systems are particularly valuable for monitoring elderly individuals and children [7, 8], enabling continuous surveillance of daily activity patterns and providing early detection of critical anomalies such as falls, irregular sleep cycles, and prolonged inactivity. Such capabilities significantly reduce the caregiving workload while improving home safety. Furthermore, these systems support chronic disease management through passive monitoring of vital parameters such as respiratory rates and heart rate variability [9].

In commercial domains, HAR technologies facilitate customer demand analysis [10], optimize operational efficiency in office environments [11], and enhance public safety by enabling real-time detection of driver fatigue and distraction in transportation systems [12, 13]. Likewise, these systems contribute to crowd density analysis, abnormal behavior detection, and infection risk assessment in public spaces [14]. Healthcare institutions, including nursing homes and hospitals, also leverage contactless monitoring to enhance caregiving efficiency, improve medical service quality, and reduce the incidence of unanticipated events through real-time vital sign detection [9, 15].

Existing non-contact sensing techniques can be broadly categorized into vision-based, IoT-based, and radiolocation-based approaches. Vision-based methods, employing far-infrared thermography [16], near-infrared gesture recognition [17], and RGBD/visible-light cameras [18–20], offer effective activity tracking but are constrained by lighting conditions, line-of-sight requirements, and privacy concerns. IoT-based techniques infer human activities indirectly through environmental changes, utilizing pressure sensors, appliance usage logs [2], and acoustic monitoring [21]. Radiolocation-based methods, including millimeter-wave radar [12, 22], ultra-wideband (UWB) systems [23, 24], and RFID-based skeleton estimation [25], achieve motion capture by analyzing electromagnetic wave propagation and reflection characteristics. However, limitations such as multipath interference, material penetration issues, and elevated hardware costs impede their large-scale, low-cost deployment.

Wi-Fi channel state information (CSI) sensing has recently attracted considerable attention for indoor HAR due to its pervasive infrastructure, fine-grained channel characterization, and minimal privacy risks. Unlike traditional Received Signal Strength Indicator (RSSI)-based techniques, CSI captures millisecond-level fluctuations in wireless channels, enabling detailed motion characterization for diverse activities—from gross motor movements like walking and running [30] to subtle physiological phenomena such as breathing patterns [32] and finger movements [31]. Its capacity for concurrent multi-user tracking further enhances its suitability for multi-member smart home environments [33], making it a practical, scalable solution for real-time, non-invasive human monitoring.

The growing adoption of edge computing has accelerated the migration of HAR systems from centralized cloud-based infrastructures to decentralized, on-device implementations. Edge computing architectures offer key advantages in latency reduction, bandwidth savings, and privacy preservation— factors particularly critical for real-time surveillance and sensitive healthcare scenarios. However, conventional Wi-Fi sensing systems rely heavily on PCs or cloud servers for model training and inference, presenting challenges related to deployment complexity, network dependency, and data privacy. While emerging studies have investigated offline Wi-Fi sensing on edge devices, most have merely adapted existing PC-based models without optimizing for the computational, memory, and energy constraints inherent to embedded edge computing platforms.

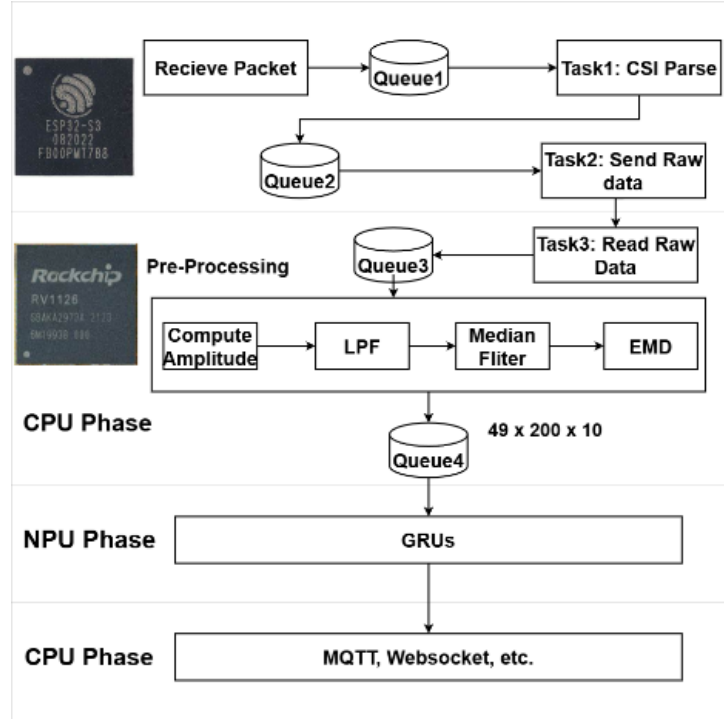

**Fig. 1.** System overview of STAR, integrating digital signal processing techniques and deep learning models for edge WiFi sensing.

To address these limitations, this paper proposes STAR (Sensing Technology for Activity Recognition), a novel Wi-Fi CSI sensing system specifically designed for low-power, resource-constrained embedded edge devices. As illustrated in Fig. 1, STAR integrates classical digital signal processing pipelines with lightweight deep learning architectures. Noise suppression and dimensionality reduction are achieved through median filtering, low-pass Butterworth filtering, and empirical mode decomposition (EMD). For inference, we introduce an attention-augmented gated recurrent unit (GRU) neural network to enhance temporal feature modeling while maintaining computational efficiency. System-level optimizations include vectorized-C code acceleration, ARM NEON instruction integration, lock-free queues for latency-critical operations, and 8-bit quantized inference to support high-throughput, low-power execution. Moreover, by offloading inference workloads to an embedded neural processing unit (NPU), STAR achieves real-time processing speeds up to 33 MHz while preserving inference accuracy.

The remainder of this article is organized as follows: Section II reviews related work in Wi-Fi-based HAR and embedded edge sensing systems. Section III details the design of the proposed system, including data acquisition, signal processing, and inference modules. Section IV presents the experimental setup, covering data collection environments, model deployment configurations, and evaluation metrics. Section V concludes with a summary of findings and directions for future research.

## II. RELATED WORKS

### A. What is the best way to process CSI data?

Deng et al. employed a Convolutional Neural Network (CNN)-based deep learning approach for Human Activity Recognition (HAR) using WiFi Channel State Information HAR via WiFi CSI signals [34]. They proposed a lightweight deep learning model, WiLDAR, which processes CSI data through a CNN after dedicated pre-processing and feature extraction stages.

In a different approach, Hernandez et al. explored federated learning (FL) for processing WiFi CSI signals [35]. Their framework, WiFederated, collaboratively trains machine learning models across multiple edge devices while safeguarding user privacy. Specifically, WiFederated enables each client to train locally on a subset of data and share only updated model parameters with a central server for aggregation, thereby reducing privacy risks associated with centralized raw data uploads.

### B. Recent Advances in CSI-Based HAR

Hernandez and Bulut [36] investigated the feasibility of WiFi sensing systems on edge devices, and analysed signal processing challenges and techniques. Through an extensive literature survey, they summarized commonly used signal processing methods and validated their effectiveness in HAR tasks of varying complexity. Their work provides a solid theoretical and practical foundation for the application of CSI in HAR.

Meanwhile, Deng et al. [34] further advanced the field by proposing WiLDAR, a lightweight model that combines stochastic convolutional kernels, depthwise separable convolutions, and residual structures. This architecture effectively reduces the network parameters and training time while achieving high accuracy, reinforcing the potential of CSI-based HAR on resource-constrained platforms.

The WiFederated framework [35] represents another significant contribution, integrating WiFi sensing with federated learning to enable collaborative model training across multiple edge locations while preserving data privacy. By avoiding raw CSI data transfer, this approach enhances model robustness and adaptability across diverse environments.

Additionally, a study in [27] introduced a methodology for highly accurate through-the-wall wireless sensing using CSI and low-cost hardware. Using a Raspberry Pi 4B with an ALFA AWUS1900 adapter, and leveraging Nexmon firmware to extract CSI data, the authors applied deep learning models based on RNN and LSTM, achieving up to 97.5% classification accuracy even in challenging non-line-of-sight (nLoS) scenarios.

Furthermore, [37] presented an innovative method applying which image processing techniques are employed to WiFi CSI-based HAR. By converting CSI data into RGB image representations and employing edge detection filters (Canny, Sobel, Prewitt, and LoG), they utilized 2D CNNs for classification of edge devices, improving both accuracy and training time. This work highlights how signal processing techniques can optimize model performance while reducing computational demands in edge deployments.

In response to these developments, we propose a lightweight architecture specifically designed for edge computing environments. Our approach reduces the number of model parameters and computational complexity, incorporates an efficient network structure, and eliminates complex pre-processing steps by enabling end-to-end learning directly from raw CSI data. This design enhances inference speed and scalability while maintaining competitive accuracy across multiple deployment scenarios, offering a practical and advanced solution for edge-based WiFi HAR.

## III. METHODOLOGY

In this section, we present the specific structure of STAR. First, we introduce the data pre-processing of the proposed network. Then, the inference method and the classification module in STAR are subsequently analysed.

### A. Data Pre-processing

In the study of WiFi signal sensing (WiFi Sensing), pre-processing and feature extraction of channel state information (CSI) data are crucial steps. These steps not only help to improve the accuracy of subsequent machine learning models, but also enhance the robustness of the system. The pre-processing flow of channel state information (CSI) is constructed through a multi-stage signal processing technique with a methodological basis derived from communication theory and a nonlinear signal processing framework.

#### 1). Calculation of CSI magnitude

Firstly, CSI data are converted into magnitude information because magnitude information is more sensitive to small changes in the physical environment. The CSI magnitude value is obtained via complex domain transformation computation, i.e., modulo the complex channel response of each subcarrier. The CSI magnitude $A(i)$ can be computed as follows:

$$|A(i)| = \sqrt{\left(h_r(i)\right)^2 + \left(h_i(i)\right)^2},$$
(1)

where $h_r(i)$ and $h_i(i)$ denote the real and imaginary parts of the $i^{th}$ subcarrier, i.e., the in-phase and quadrature components, respectively. This step is based on the fact that the amplitude property of the complex signal makes it effective in retaining the main energy information of the signal while removing the possible noise in the phase information. The amplitude information is more robust to multipath effects and hardware noise than the phase information is.

#### 2). Median Filtering

Median filtering is performed on the CSI amplitude data to remove outliers in the data. The median filtering stage uses a nonlinear sliding window algorithm to replace the value of each sampling point with the median of the data in the neighbourhood, and selects the median value within the data window as the output, thus effectively suppressing noise and impulse interference while preserving the sharpness characteristics of the action signal. For a sliding window of length $w=2k+1$ ($k$ is the window radius), the CSI amplitude signal is a discrete time series $x(n)$, and the median filtered output $y(n)$ can be expressed as:

$$y(n) = \text{median}\{x(m)|m \in [n-k, n+k]\},$$
(2)

where the *median* {} denotes the median-taking operation. Boundary processing at the beginning and end of the signal is performed by reducing the window length, which is {0, $n+k$} if $n<k$. Median filtering is chosen for its properties with respect to the preservation of information at the edges of the signal, avoiding the blurring effect of mean filtering, as well as for its effective rejection of non-Gaussian noise (e.g., bursts of interference).

### 3). Low-pass filtering via an $8^{th}$-order Butterworth filter

In this step, we process the signal via an $8^{th}$-order Butterworth low-pass filter, which is used to further smooth the CSI signal and remove the high-frequency noise. Butterworth filters are known for their flatness of frequency response in the passband and are suitable for application scenarios where the low-frequency component of the signal needs to be maintained. For an 8 -order Butterworth filter, the amplitude squared response equation is in Eqn. 3: the magnitude squared function of the Butterworth filter characterizes how the filter attenuates frequencies relative to the cut-off frequency $\omega_c$, with the higher the order, the steeper the attenuation outside the passband.

$$|H(j\omega)|^2 = \frac{1}{1 + \left(\frac{\omega}{\omega c}\right)^{16}},$$

(3)

where, $\omega$: is the angular frequency (rad/s), and $\omega_c$: is the cut-off angular frequency that determines the demarcation between the passband and stopband.$16=2\times N$, where $N=8$; the higher the filter order $N$ is, the steeper the transition band. In the passband ($\omega<\omega_c$), the amplitude response is as flat as possible; in the stopband ($\omega>\omega_c$), the amplitude decays rapidly. The poles of a Butterworth filter lie on the unit circle of the complex plane and are uniformly distributed. For an $8^{th}$-order filter, the pole equation is:

$$p_k = e^{j\left(\frac{\pi}{2} + \frac{(2k+1)\pi}{2N}\right)}, k = 0,1, \ldots, 7,$$

(4)

where $p_k$: complex coordinates of the $k^{th}$ pole, $j$: imaginary unit, and the poles are uniformly distributed on the unit circle to ensure that the amplitude response is maximally flat in the passband. The normalized low-pass filter design steps are divided into: normalizing as follows: normalize the cut-off frequency: make $\omega_c=1$, and design the prototype filter. Denormalization: mapping the prototype filter to the actual cut-off frequency via frequency transformation $\omega_c$. The bilinear transformation compensates for the nonlinearity of the frequency response by mapping the analogue frequency to the digital frequency through nonlinear frequency compression, which serves to convert the analogue transfer function $H(s)$ to the digital transfer function $H(z)$:

$$s = \frac{2}{T} \cdot \frac{z-1}{z+1},$$

(5)

$T$: sampling period (related to the CSI signal sampling rate), $z$: digital domain complex variable ($z=e^{j\omega}$). The transfer function of the final digital filter is as follows:

$$H(z) = \frac{\sum_{k=0}^{8} b_k z^{-k}}{1 + \sum_{k=1}^{8} a_k z^{-k}}.$$

(6)

The numerator is calculated via the *scipy.signal.butter* function ($b_k$), and the denominator ($a_k$) coefficients are calculated.

Real-time computation is achieved by processing the CSI amplitude signal with difference equations, where each output sample $y[n]$ is obtained via weighted summation of the current input $x[n]$ and the historical inputs and outputs. The difference equation of the filter is given by:

$$y[n] = \sum_{k=0}^{8} b_k x[n-k] - \sum_{k=1}^{8} a_k y[n-k].$$

(7)

Each output sample $y[n]$ is obtained via weighted summation of the current input $x[n]$ and historical inputs and outputs. $x[n-k]$: past value of the input signal. $y[n-k]$: past value of the output signal

The frequency response needs to be verified after design:

$$H(e^{j\omega}) = \frac{\sum_{k=0}^{8} b_k e^{-j\omega k}}{1 + \sum_{k=1}^{8} a_k e^{-j\omega k}},$$

(8)

$\omega$: digital corner frequency (range: $0\leq\omega<\pi$). Usually, a Butterworth filter passband without ripple is used. Stopband attenuation ($8^{th}$-order provides at least 48dB/decade attenuation).

### 4). EMD decomposition to remove high -frequency components

The empirical modal decomposition (EMD) technique is employed in the context of competing for nonlinear non-smoothed signal features. This technique involves the decomposition of the CSI signal to eliminate high-frequency components. EMD is a method of signal decomposition method that is adaptive in nature. It is capable of decomposing a nonlinear, non-smoothed signal into a series of intrinsic modal functions (IMFs). Each IMF represents fluctuations in the signal at different frequency scales. The removal of high-frequency IMF components allows for the subsequent smoothing of the CSI signal, thereby enhancing the efficacy of feature extraction. The process of EMD decomposition can be delineated as follows:

$$x(t) = \sum_{i=1}^{n} \text{IMF}_i(t) + r_n(t),$$

(9)

where $x(t)$ is the original signal, IMF$i(t)$ is the $i^{th}$-order high-frequency to low-frequency Eigen mode function, and $r</a6>n(t)$ is the residual signal. The process of signal reconstruction with its core equation

$$x_{\text{filtered}}(t) = \sum_{i=k}^{n} \text{IMF}_i(t) + r_n(t).$$

(10)

where $k$ is the retained IMF starting order, and the filtering intensity is directly controlled by adjusting the value of $k$

without complex parameter settings. In practice, high-frequency noise is usually located in the first few orders, and you can choose to remove the first few high-frequency IMF components can be removed as needed to smooth the signal. The core of EMD decomposition is to decompose the signal into IMF components through a "filtering" process. Although there is no explicit mathematical formula for EMD decomposition, the filtering process can be summarized in the following steps: First, all local maxima and minima of the signal are identified. Then the upper and lower envelopes are subsequently constructed using cubic spline interpolation. The mean value of the upper and lower envelopes is then calculated and subtracted from the original signal to obtain a new signal. The above steps are repeated until the stopping criterion of the IMF is satisfied. Among the IMF components obtained by EMD decomposition, the high-frequency component usually contains noise information, whereas the low-frequency component contains valid signal features.

*5). Normalization*

Finally, the pre-processed CSI data are normalized to ensure that the input features have a uniform scale. Normalization is an important step to improve the efficiency and performance of machine learning model training, which can improve the convergence speed and performance of the machine learning model and make sure that different features have the same weight in model training. Commonly used normalization methods include min-–max normalization and *Z* Z-score normalization. We use Min-Max normalization:

$$x_{\text{norm}} = norm \frac{x - x_{min}}{x_{max} - x_{min}},$$

(11)

where $x_{min}$ and $x_{max}$ are the minimum and maximum values of the signal, respectively. This operation removes the magnitude difference so that the data distribution satisfies $x_{\text{norm}} \in [0, 1 </a9>]$, which conforms to the distributional assumptions that most machine learning algorithms make about the input data.

By performing magnitude calculation, median filtering, low-pass filtering (using an $8^{th}$-order Butterworth filter), and EMD decomposition to remove high-frequency components, and normalization on the CSI data to form perform complete CSI pre-processing and feature extraction process, we obtain the denoised CSI magnitude values. To reduce the computational complexity and improve the model performance, the amplitude values of the first 49 subcarriers are chosen to be retained as feature inputs to the subsequent machine learning model. These features not only remove the noise and high-frequency components, but also retain the key information in the signal, providing effective inputs for subsequent activity recognition or localization tasks.

*B. Inferences Method*

The gated recurrent unit (GRU) is a recurrent neural network variant proposed by Cho et al. [38][35], aiming that aims to solve the vanishing gradient problem of traditional RNNRNNs in long -sequence modelling. As shown in Fig. 2, the GRU effectively regulates the information flow by introducing a gating mechanism, which significantly reduces the computational complexity while preserving the model expressiveness. By simplifying the LSTM structure and merging three gates (input gate, forget gate, and output gate) into two gates (update gate and reset gate), the number of parameters and computational complexity are reduced, while retaining the ability to model long -term dependencies, thus providing a more advantageous computational efficiency while still maintaining a high prediction accuracy.

The GRU structure consists of two primary gates: the update gate ($z_t$, red) controls how much previous memory to retain; the reset gate ($r_t$, blue) determines how much past information to forget. The candidate hidden state ($\sim h_t$, green) computes new memory content. Element-wise operations ($\odot$, 1-, +) facilitate information flow control. This simplified architecture requires fewer parameters than LSTM while maintaining comparable performance on sequential data tasks.

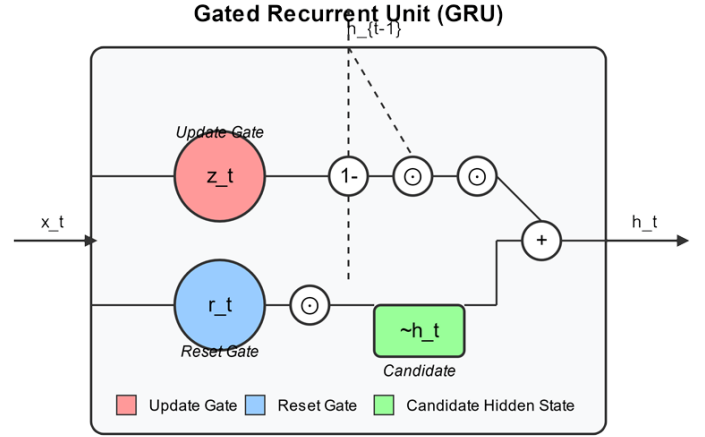

**Fig. 2.** GRU cell architecture.

**Algorithm 1: GRU Forward Propagation**

**Input:** Input sequence X = {x₁, x₂, ..., x_T},
parameter matrices W_z, W_r, W
and biases b_z, b_r, b

**Output:** Hidden state sequence H = {h₁, h₂, ..., h_T}

**Initialize:** h₀ = 0 (zero vector)

**for** t from 1 to T **do**

  1: Compute update gate:
    z_t = σ(W_z·[h_{t-1}, x_t] + b_z)

  2: Compute reset gate:
    r_t = σ(W_r·[h_{t-1}, x_t] + b_r)

  3: Compute candidate hidden state:
    h̃_t = tanh(W·[r_t⊙h_{t-1}, x_t] + b)

  4: Update hidden state:
    h_t = (1-z_t)⊙h_{t-1} + z_t⊙h̃_t

**end for**

**return** H

**Fig. 3.** Algorithmic Pseudo-Code.

The GRU has multiple advantages over traditional RNNs and LSTM. First, the simplified gating mechanism drastically reduces the number of parameters and its reduces the computational complexity by approximately 33% without significantly affecting the model performance [39][36]. Empirical studies have shown that the GRU performs comparably to the LSTM in a wide range of sequence modelling tasks, but tends to perform better in small datasets and tasks that require capturing short-term dependencies [40][37]. In addition, the GRU has shorter computational paths and smoother gradient flow, effectively mitigating the gradient vanishing gradient problem [41][38].

Empirical studies have shown that GRUs perform well in a wide range of sequence modelling tasks. A large-scale evaluation of more than 10,000 RNN architectures by Jozefowicz et al. [39][36] showed revealed that the performance of GRUs in language modelling tasks is comparable to that of more complex models. In terms of computational efficiency, a study by Britz et al. [42][39] found revealed that the GRU is, on average about, approximately 25% faster than the LSTM in training time, and is only about 1-2%approximately 1–2% less accurate in neural machine translation tasks.

The GRU model effectively solves the long-term dependency problem in traditional RNNs through its unique gating mechanism and achieves excellent performance in a variety of sequence modelling tasks. Its simplified structure not only reduces computational complexity, but also alleviates the gradient vanishing problem, making it an efficient choice for processing sequence data. With the development of deep learning, the GRU has become an important basic model for time series data processing, providing strong support for many cutting-edge applications. Fig. 3 shows the operation in pseudocode.

## IV. EXPERIMENTATION

### A. Data Acquisition

#### 1) Acquisition device

In this study, the ESP32-S3 Dongle (Fig. 4) is employed as the CSI data acquisition platform. ESP32-S3 is a low-power, compact microcontroller (MCU) that integrates a WiFi RF transceiver, is compatible with the IEEE 802.11n standard and supports both 20 MHz and 40 MHz bandwidth configurations. It offers an accessible C/C++ development environment alongside dedicated CSI data acquisition libraries.

For data collection, two ESP32-S3 Dongles with different firmware configurations are utilized: one operating as a transmitter and the other operating as a receiver. Upon activation, the transmitter autonomously initiates the transmission of empty data packets. Concurrently, the receiver establishes a connection with the transmitter and enters an active state, receiving these packets and extracting the associated CSI feature information. The receiver then communicates with a PC via a serial port (Fig. 5), transmitting the CSI feature data at a rate of 100 frames per second. A dedicated data capture application on the PC records the CSI data in a LevelDB database for subsequent processing.

The data capture software (Fig. 6) systematically stores raw CSI data at a frequency of 100 entries per second. Each entry comprises a timestamp alongside auxiliary metadata such as signal strength and device MAC addresses. However, this study exclusively focuses on the CSI features themselves. These features are stored sequentially in subcarrier order, preserving both the real and imaginary coefficients for each CSI entry. Additionally, the capture software interface enables real-time monitoring of sampling values during the data acquisition process.

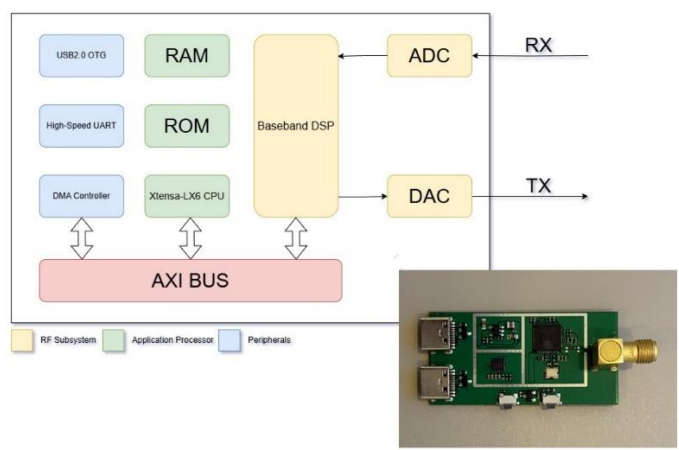

**Fig. 4.** ESP32-S3 Dongle.

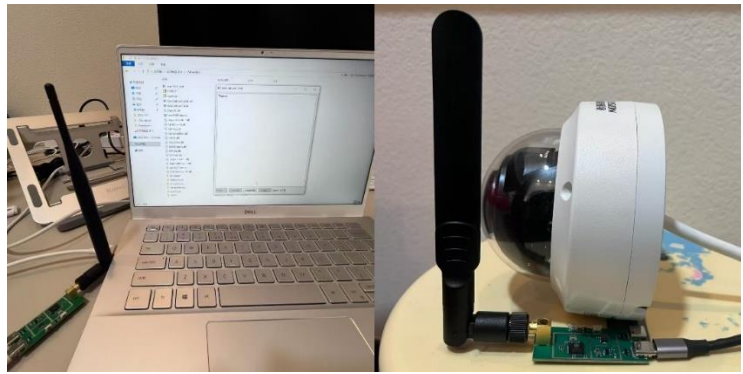

**Fig. 5.** Connection diagram illustrating the setup of the transmitter, receiver, and PC.

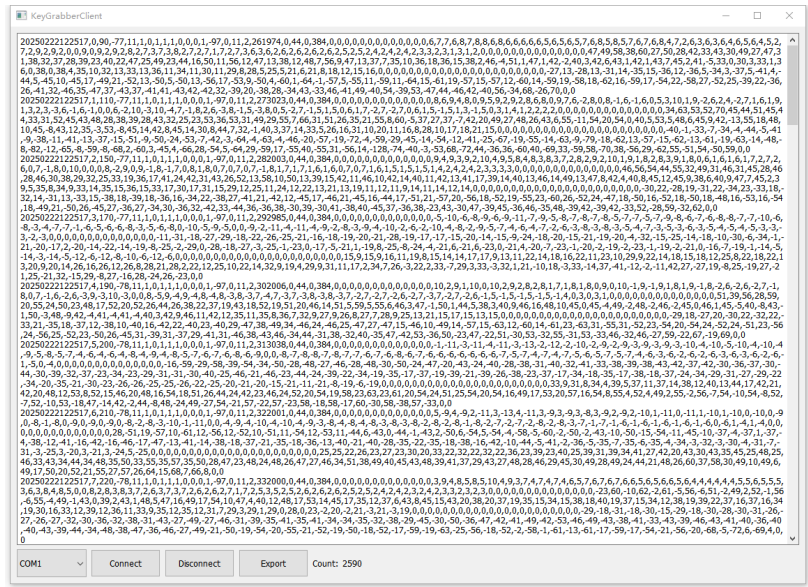

**Fig. 6.** User interface of the data capture software displaying the real-time data logs being recorded.

*2). Data Acquisition Environment*

A relatively empty room was selected for the experiment, where the transmitter and receiver were positioned at different locations. A camera was used to continuously record the scene, with time stamps on the video to help identify and differentiate various actions. The volunteers performed several distinct actions between the transmitter and receiver, as illustrated in Fig. 7.

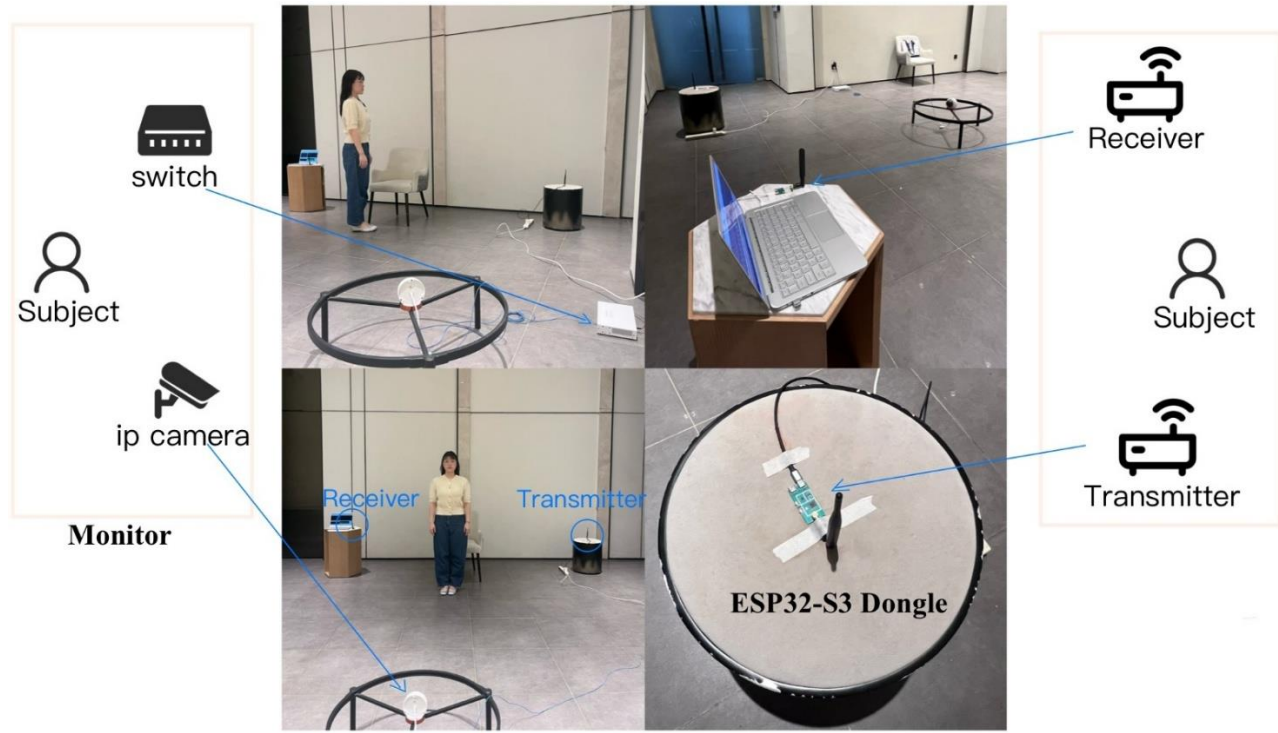

**Fig. 7.** Live view of the data acquisition environment, showing the positioning of the transmitter, receiver, and volunteer performing actions.

*3). Data overview*

The parameters of the receiver side of the collector are configured as follows:

```
.lltf_en = true,
.stbc_htltf_en = false, .stbc_htltf2_en = false, .
.stbc_htltf2_en = false,
.ltf_merge_en = true, .
.channel_filter_en = false, .
.channel_filter_en = false, .manu_scale = false, .
.shift = false,
```

In this experiment, the Legacy Long Training Fieldlegacy long training field (LLTF) is activated, with the filter disabled. The channel bandwidth is set to 20 MHz, and each transmission of CSI data contains 52 subcarriers, each organized as "real-virtual" coefficients, resulting in 104 coefficients per data group.

The data are transmitted at a frequency of 100 Hz, meaning that a new set of data is sent every 10 ms.

Volunteers perform seven different postures: lying down, falling, walking, picking up, running, sitting down, and standing up, along with a "no one" state, totalling 200,000 data groups. The entire dataset has a size of 104×200,000×200000.

*B. Data Pre-processing*

The pre-processing of the CSI data begins by calculating the amplitude values of the CSI data. Afterward, both median filtering and low-pass filtering are applied to the amplitude data using an 8th-order Butterworth filter to remove noise and smooth the signal. Finally, empirical mode decomposition (EMD) is employed to extract and remove high-frequency components from the data, enhancing the signal quality for further analysis.

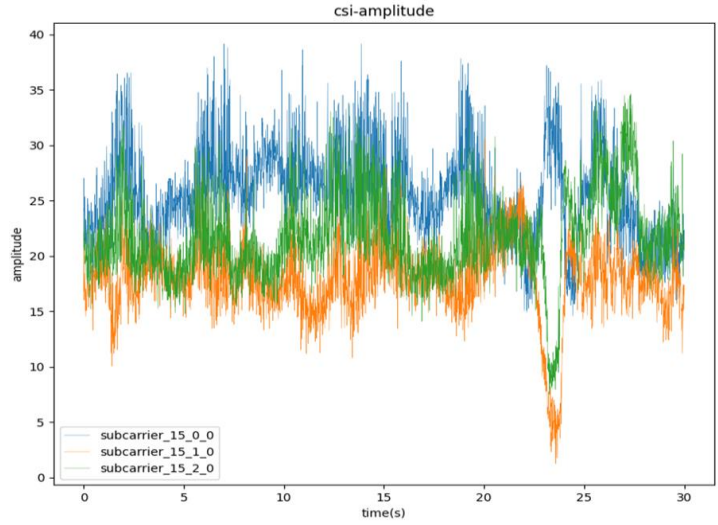

**Fig. 8.** Amplitude of the CSI data before pre-processing.

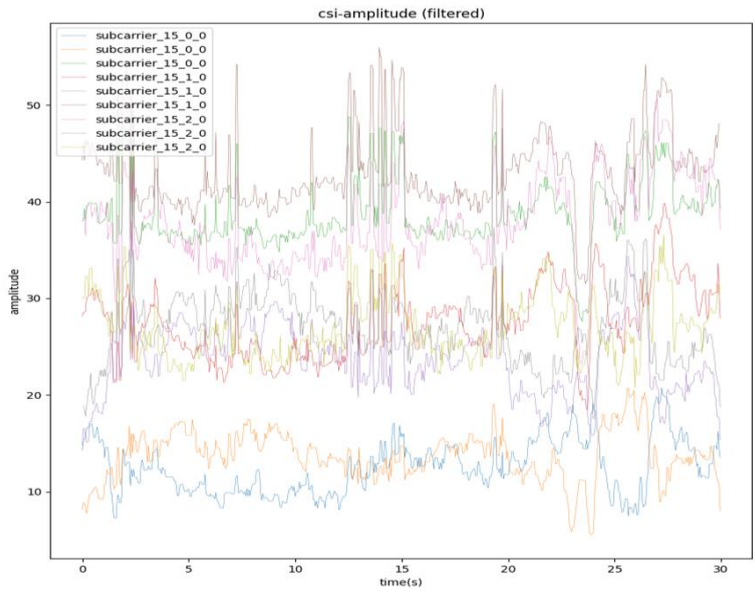

**Fig. 9.** Amplitude of the CSI data after median and low-pass filtering.

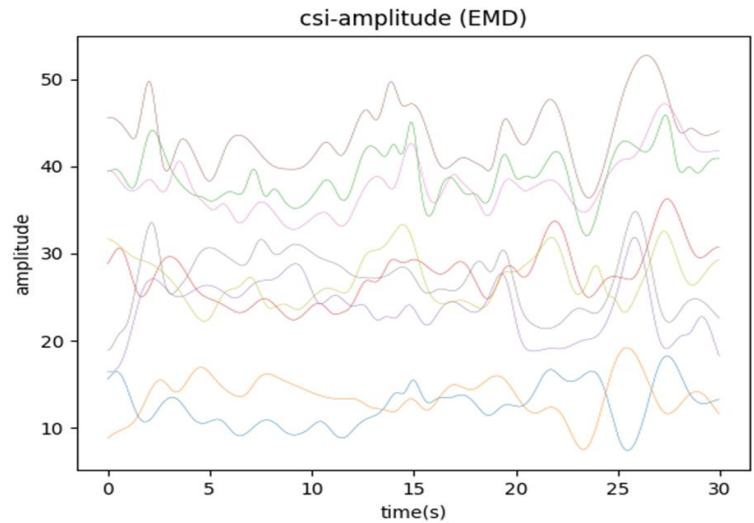

**Fig. 10.** Amplitude of the CSI data after high-frequency component removal using EMD.

### C. Model Training

#### 1). Training dataset preparation

The pre-processed training set consists of 49 features, derived from 160,000 out of 200,000 sets of data. The data are grouped into 200 samples per batch to ensure that all actions within each group are consistent. As a result, the training set has dimensions of 49×200×80049, corresponding to 7 gesture categories and two states (manned and unmanned). All the data are labelled appropriately and undergoes normalization. The remaining data are grouped similarly and used as the test set.

#### 2). network structure

As illustrated in Fig. 11, the network uses a 3-layer GRU architecture within a recurrent neural network (RNN). To enhance the system's robustness, the detection of occupied and unoccupied states is incorporated. This helps distinguish noise signals from action signals during unoccupied states, reducing the risk of false alarms in action classification. By replacing LSTM with a GRU, the network structure is simplified, which lowers computational demands and facilitates easier deployment on edge devices. Cross-entropy is used as the loss function for both person presence detection and action classification.

#### 3). Model Training

The training framework is implemented using PyTorch. The batch size is set to 200, representing the group size, with 200 iterations in total. The model achieves an average classification accuracy of 93.52% across 80,000 test samples. The classification accuracy for each category is shown in TABLE I.

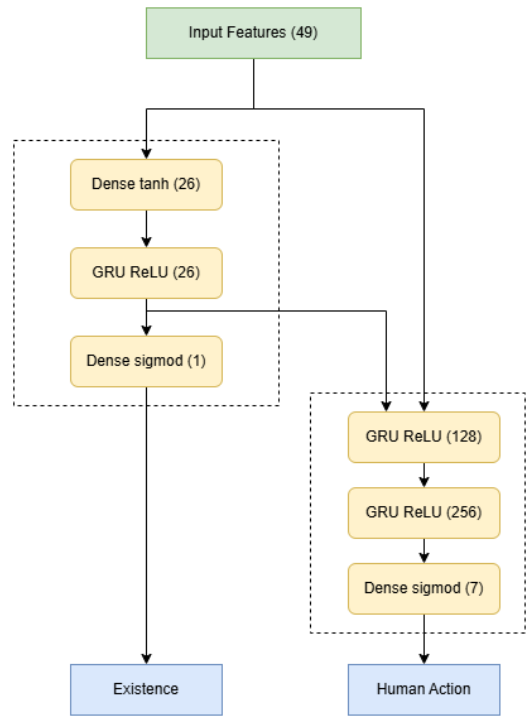

**Fig. 11.** Network topology of STAR, featuring a multi-layer GRU architecture for action classification and person presence detection.

### D. Edge Computing

#### 1). Hardware Platform

Traditional CSI sensing systems typically rely on the computational resources of PCs or the cloud infrastructure, which introduce significant concerns regarding data privacy, deployment complexity, and associated costs. To address these challenges, we propose an offline deployment method that utilizes portable hardware devices for model inference, eliminating the dependency on PCs and cloud services. This approach enhances the practicality of WiFi sensing technologies in real-world engineering applications.

Recent studies have explored the application of edge computing to WiFi sensing, with implementations using MCU- or CPU-based platforms such as Raspberry Pi and ESP32, as well as GPU-based solutions such as the Jetson series of single-board computers. However, we argue that the computational power of the former is insufficient to ensure real-time data processing while maintaining inference accuracy. In contrast, while the latter provides substantial computational power, its large size and high power consumption hinder its deployment in field applications. Furthermore, both the Raspberry Pi and Jetson devices lack integrated CSI data capture modules, necessitating the use of external receivers, which further complicates deployment.

In our experiments, we utilize a custom-built hardware platform based on the Rockchip RV1126 processor, which integrates the ESP32-S3, allowing for seamless CSI data acquisition and processing within a single device. A key feature of the RV1126 is its internal neural processing unit (NPU), which is capable of 2 TOPS (tera operations per second), enabling real-time inference without overburdening the CPU (Fig. 12). This integration significantly improves the efficiency and convenience of deploying WiFi sensing technology in practical applications.

TABLE I
MODEL CLASSIFICATION ACCURACY

| Class of activity | Accuracy |
| --- | --- |
| lie down | 96.12% |
| fall | 85.22% |
| walk | 90.11% |
| pickup | 94.55% |
| run | 88.71% |
| sit down | 96.90% |
| stand up | 97.46% |
| Have person or No person | 99.11% |

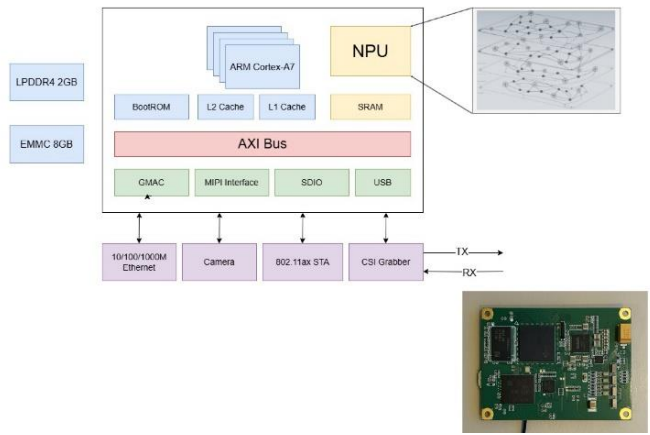

**Fig. 12.** RV1126-based edge hardware.

### 2). Performance Evaluation

The proposed network comprises 97,616 weight parameters, with the core pre-processing and RNN inference computations implemented in vectorized C. The performance evaluation of our network is summarized in Table II.

In CPU mode, the performance is assessed under both the FP16 and INT8 quantization formats, using a 1.5 GHz Cortex-A7 processor integrated within the RV1126 platform. When switching to the NPU mode, the model undergoes INT8 quantization via Rockchip's toolchains, resulting in a test inference speed of 33 MHz. This represents a six-fold increase in performance compared to that of the CPU mode. Furthermore, in the NPU mode, the CPU is solely engaged in data pre-processing, reducing its occupancy to 8%. Notably, the use of INT8 quantization does not significantly impact the inference accuracy, thereby balancing enhanced computational efficiency with minimal loss in performance.

TABLE II
PERFORMANCE EVAULATIONEVALUATION IN FP16
AND INT8 QUANTIZATION MODES

| Quantization Accuracy | Required Arithmetic Power | Reasoning Speed | CPU Occupancy |
|---|---|---|---|
| INT8 | 48Mflops | 5000KHz | 28% |
| FP16 | 166Mflops | 1800KHz | 56% |

## V. CONCLUSION

This study introduces a novel approach for Wi-Fi CSI-based human activity recognition (HAR) by leveraging a 3-layer GRU-based recurrent neural network (RNN) architecture. The proposed model achieves high classification accuracy with minimal computational complexity, making it well- suited for deployment on resource-constrained edge devices. To meet the real-time inference demands, we implemented core computations via vectorised C code, enabling the RV1126 platform to maintain a CSI sampling rate exceeding 100Hz100 Hz, ensuring efficient data processing and reliable performance in practical environments.

Additionally, NPU (neural processing unit) acceleration plays a crucial role in enhancing inference speed, achieving up to six times the processing rate of CPU-based execution. Offloading inference to the NPU significantly reduces CPU utilization to just 8%, allowing the CPU to focus on other tasks, such as data pre-processing. Notably, the adoption of INT8 quantization during model deployment did not compromise accuracy, demonstrating the effectiveness of quantization in improving computational efficiency without sacrificing performance.

The proposed system integrates lightweight modelling, hardware optimization, and adaptive signal processing techniques to offer an efficient solution for real-time single-receiver Wi-Fi sensing. This work establishes a robust, privacy-preserving edge AI framework, capable of performing accurate activity recognition without relying on cloud services, thus addressing concerns related to data privacy and scalability in real-world deployments.

By combining efficient model design, hardware acceleration, and low-power deployment, our system represents a significant advancement in the practical application of Wi-Fi CSI sensing for edge AI. It lays the foundation for diverse use cases in smart home monitoring, healthcare, and security systems, offering a reliable and energy-efficient solution for real-time activity recognition in privacy-conscious environments.